\documentclass[11pt]{article}
\usepackage[preprint]{acl}
\usepackage{times}
\usepackage{latexsym}
\usepackage[T1]{fontenc}
\usepackage[utf8]{inputenc}
\usepackage{microtype}
\usepackage{inconsolata}
\usepackage{graphicx}

\usepackage{booktabs}
\usepackage{tcolorbox}
\usepackage{multirow}
\usepackage[table]{xcolor}
\usepackage{tikz}
\usepackage{newunicodechar}
\newunicodechar{↗}{\textsuperscript{$\nearrow$}}

\newcommand{\red}[1]{\textcolor{red!80!black}{#1}}
\newcommand{\green}[1]{\textcolor{green!70!black}{#1}}
\title{\textsc{NarraBench}:\\ A Comprehensive Framework for Narrative Benchmarking}

\author{Sil Hamilton \\ Cornell University
        \And 
        Matthew Wilkens \\ Cornell University
        \And
        Andrew Piper \\ McGill University}

\begin{document}
\maketitle
\begin{abstract}
 We present \textsc{NarraBench}, a theory-informed taxonomy of narrative-understanding tasks, as well as an associated survey of 78 existing benchmarks in the area. We find significant need for new evaluations covering aspects of narrative understanding that are either overlooked in current work or are poorly aligned with existing metrics. Specifically, we estimate that only 27\% of narrative tasks are well captured by existing benchmarks, and we note that some areas -- including narrative events, style, perspective, and revelation -- are nearly absent from current evaluations. We also note the need for increased development of benchmarks capable of assessing constitutively subjective and perspectival aspects of narrative, that is, aspects for which there is generally no single correct answer. Our taxonomy, survey, and methodology are of value to NLP researchers seeking to test LLM narrative understanding. 
\end{abstract}

\section{Introduction}

Narratives are omnipresent in daily life. 
They can be used to entertain, inform, persuade, and maintain shared beliefs in communities across generations. 
Designing large language models (LLMs) that can both understand and generate narrative communication is of utmost importance to building artificially intelligent systems useful for humans.

\begin{figure}[t]
\centering
\includegraphics[width=0.88\linewidth]{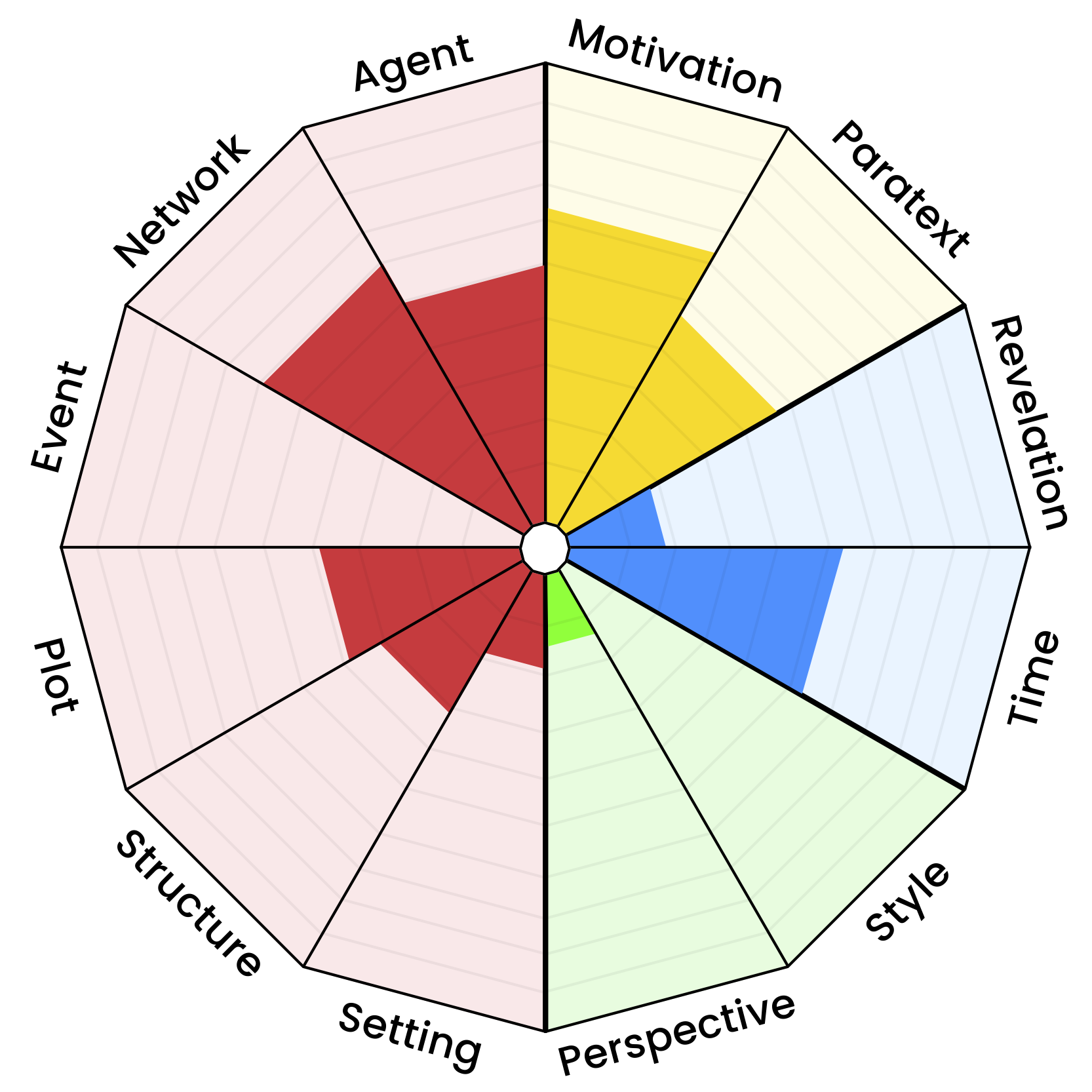}
\caption{The twelve primary narrative features of the \textsc{NarraBench} taxonomy coloured according to the Big-4 narrative dimensions (\textcolor[HTML]{E4383B}{story}, \textcolor[HTML]{11FF00}{narration}, \textcolor[HTML]{0090FF}{discourse}, and \textcolor[HTML]{FFDB00}{situatedness}) and shaded by how well existing benchmarks match as determined by our survey.}
\label{fig:benchmark-distributions}
\end{figure}

Using narratives to test long context understanding and models' theory of mind has come into vogue of late \citep{kimFANToMBenchmarkStresstesting2023,karpinskaOneThousandOne2024,chenMBENCHBenchmarkingTheory,wangNovelQABenchmarkingQuestion2025}, and non-narrative-specific benchmarks can test for narrative understanding in unexpected ways: ``Needle in a haystack'' tests show models can discern what is relevant to the present contextual frame \citep{kamradtNeedleHaystackPressure2025}, and summarization tasks require models to integrate information across long contexts to identify important through-lines \citep{linROUGEPackageAutomatic2004}.

Several large-scale benchmarks contain a significant amount of narrative content in this way.
For example, we estimate approximately 1 in 7 BIG-bench questions contains a story \citep{srivastavaImitationGameQuantifying2023}, 1 in 4 in HellaSwag \citep{zellersHellaSwagCanMachine2019}, and 1 in 3 in MMLU \citep{hendrycksMeasuringMassiveMultitask2021}.\footnote{All datasets were classified on a per-document basis with the BERT-based classifier from \citet{antoniakWherePeopleTell2024}.}
Correctly answering these questions indicates some level of narrative understanding.

Nevertheless, benchmarks that \emph{explicitly} and \emph{systematically} test for narrative understanding are rare and often lack theoretical consistency in their principal goals and methodological foundations.
Existing benchmarks often deploy single-answer frameworks, overlooking the way narrative understanding is a fundamentally interpretive (i.e., perspectival, not deterministic) task.
Benchmarks likewise tend to focus on story-level content, missing more complex dimensions of narrative communication (e.g. discourse structure and narrative perspective).

In this paper, we propose a novel comprehensive benchmarking framework designed to assess LLMs' capacity for narrative understanding. 
Our framework incorporates relevant existing benchmarks and identifies areas for future development under a single, theoretically coherent umbrella. 
Our core contributions are: 

\paragraph{Survey of existing benchmarks for narrative understanding.} 
We review 78 existing benchmarks for their relevance for the task of narrative understanding. 
Of these, we find that 39 lack available data, while the remaining 39 offer reasonable or good fit with key aspects of narrative understanding. Of the 39 plausible and available benchmarks, 32 still fail to align exactly with narrative theory.
As can be seen in \autoref{fig:benchmark-distributions}, existing benchmarks cover approximately 27\% of our taxonomy.
We highlight the consistent focus on deterministic, story-level understanding tasks as a major limitation to existing benchmark design.
We finally describe a method for implementing benchmarks in a unified testing harness to serve as a reference implementation of the \textsc{NarraBench} framework.

\paragraph{A novel taxonomy of narrative understanding tasks.} 
We develop a taxonomy of fifty distinct narrative understanding tasks derived from well-established theoretical frameworks in narratology and state of the art benchmarking theory.
This taxonomy provides the first systematic integration of narrative understanding benchmarks within a unified theoretical framework that allows for future expansion. 
It also foregrounds the importance of perspectival alignment in benchmark development.

\paragraph{Charting a path forward.} 
We highlight where existing benchmarks fit within this new taxonomy and where there are areas for future work. 
Our taxonomy provides a usable, expandable road-map for the development of new benchmarks and benchmarking data to create a robust resource for assessing LLM performance on a central aspect of human communication and cultural behaviour.

\section{Related Work}
\label{sec:relatedwork}

\paragraph{Narrative Understanding.} Early computational work modelled narratives as stochastic processes \citep{kahnFinitestateModelsPlot1973} or scripts known as \emph{Fillmorean frames} \citep{schankScriptsPlansGoals1975} containing chronological chains of events composed of entities and their actions \citep{chambersUnsupervisedLearningNarrative2008,chambersUnsupervisedLearningNarrative2009,chambersDatabaseNarrativeSchemas2010,balasubramanianGeneratingCoherentEvent2013,reiterNLPbasedCrossdocumentApproach2014}.
% script extraction (middle)
Advancements in NLP led to increasingly sophisticated narrative extraction methods that made use of topic models, manually derived features, and associated rule mining
\citep{antoniakNarrativePathsNegotiation2019a,belyyScriptInductionAssociation2020,lyuGoalOrientedScriptConstruction2021,zhangHumanintheLoopSchemaInduction2023}.
This approach was extended to plot arc extraction, newly reinterpreted as sentiment over time \citep{hoganAffectiveNarratologyEmotional2011,reaganEmotionalArcsStories2016,somasundaranEmotionArcsStudent2020,elkinsShapesStoriesSentiment2022, knight2024narrative}, and social network analysis, where entities and their actions were rendered as graphs \citep{roemmeleIdentifyingSensibleLexical2019,simsMeasuringInformationPropagation2020a, tangherlini2020automated}. Further work has focused on higher-level understanding including narrativity detection \cite{antoniakWherePeopleTell2025, piper-bagga-2025-narradetect} and narrative intent understanding \cite{zhuAreNLPModels2023}. As LLMs are increasingly used to assess and generate narratives, now is an optimal time to codify the diverse array of tasks that fall under the heading of narrative understanding. 

\paragraph{Advances and limitations in current benchmarking.}

Benchmarks are the principal infrastructure of empirical NLP, enabling standardized comparison and shared progress measures across models.
Of particular interest has been benchmarks for evaluating long-context understanding \citep{changBooookScoreSystematicExploration2024,karpinskaOneThousandOne2024,wangNovelQABenchmarkingQuestion2025,thaiLiteraryEvidenceRetrieval2025} and summarization performance \citep{fabbriSummEvalReevaluatingSummarization2021,zhaoNarraSumLargeScaleDataset2022,subbiahSTORYSUMMEvaluatingFaithfulness2025}.
A smaller set of benchmarks focus on explicitly narrative phenomena---including dialogue \citep{vishnubhotlaProjectDialogismNovel2022}, plot arcs \citep{chunSentimentArcsNovelMethod2021a}, or genre classification \citep{worshamGenreIdentificationCompositional}.
General benchmarks for narrative understanding remain scarce.
Evaluation methods also remain limited; most rely on multiple-choice questions for efficiency --- increasingly seen as unreliable estimates of model comprehension \citep{sapSocialIQaCommonsense2019,wangMyAnswerFirstToken2024a,balepurArtifactsAbductionHow2024,alzahraniWhenBenchmarksAre2024,wangLLMsMayPerform2025}.
Recent alternatives such as LLM-as-a-Judge evaluation \citep{liLLMsasJudgesComprehensiveSurvey2024,tanJudgeBenchBenchmarkEvaluating2025,wangImprovingLLMasaJudgeInference2025} and distributional or multi-annotator scoring \citep{niMixevalDerivingWisdom2024,meisterBenchmarkingDistributionalAlignment2024} point toward richer, less rigid evaluation pipelines, but have not yet been applied to narrative domains. 

\section{Theoretical Foundations}
Narrative theory has developed, over more than a century, increasingly sophisticated qualitative descriptions of narratives across different mediums and contextual settings.

\subsection{The Elements of Narrative}

Our base definition of narrative communication is drawn from \citet{piperNarrativeTheoryComputational2021}. At its most elementary level, a narrative can be said to occur when all of the following criteria are met:

\begin{table}[h!]
\begin{center}
\begin{tabular}{cc}
A&Someone \\ 
B&tells\\ 
C&someone \\
D&somewhere \\
E&for some reason \\
\hline  
 &that \\
\hline
F&someone\\ 
G&did something(s)\\ 
H&[to/with someone]\\ 
I&somewhere\\ 
J&at some time\\
K&for some reason.\\ 
\end{tabular}
\end{center}
\vspace{-1em}
\end{table}

For there to be a narrative, we need (A) a teller, (B) a mode of telling (i.e., medium), (C) a recipient, (D) a social situation, (E) a motivation or intent for telling the story, (F) an agent, (G) at least one action or event, (H) a possible object, (I) a location, (J) a time-frame, and (K) a motivation or cause of the actions involved. 
Narratologists distinguish between the frame of the storyworld (the elements that come after the double lines above) known as ``diegetic'' elements, and the frame of telling (those elements that come before the double lines) known as ``heterodiegetic'' elements, where diegesis refers to a narrative ``frame'' or ``world.'' These dimensions need not be explicit, but they do need to be implied for a coherent narrative to form.

Narrative theory has additionally foregrounded the centrality of change or conflict as an elementary component of narrative communication \cite{herman2009basic}, through concepts like ``change of state'' \cite{prince_narratology_2012}, ``canonicity/breach'' \cite{bruner_narrative_1991}, denouement \cite{freytag1895technique}, or Aristotle's foundational emphasis in the \emph{Poetics} (c. 330 \textsc{bce}) on beginning, middle, and end structure. 
At a root level, narratives foreground \emph{temporal difference} with respect to agent-centred events \cite{genetteNarrativeDiscourseEssay1980, sternberg1992telling, ricoeur2012time}. 

\subsection{Higher-Order Narrative Dimensions}

\begin{figure}[t]
\centering
\begin{tcolorbox}[colframe=black,boxrule=0.7pt,sharp corners,
                  left=3pt,right=3pt,top=3pt,bottom=3pt,boxsep=3pt,arc=3pt]
\centering
\includegraphics[width=0.85\linewidth,trim={12pt 12pt 12pt 12pt},clip]{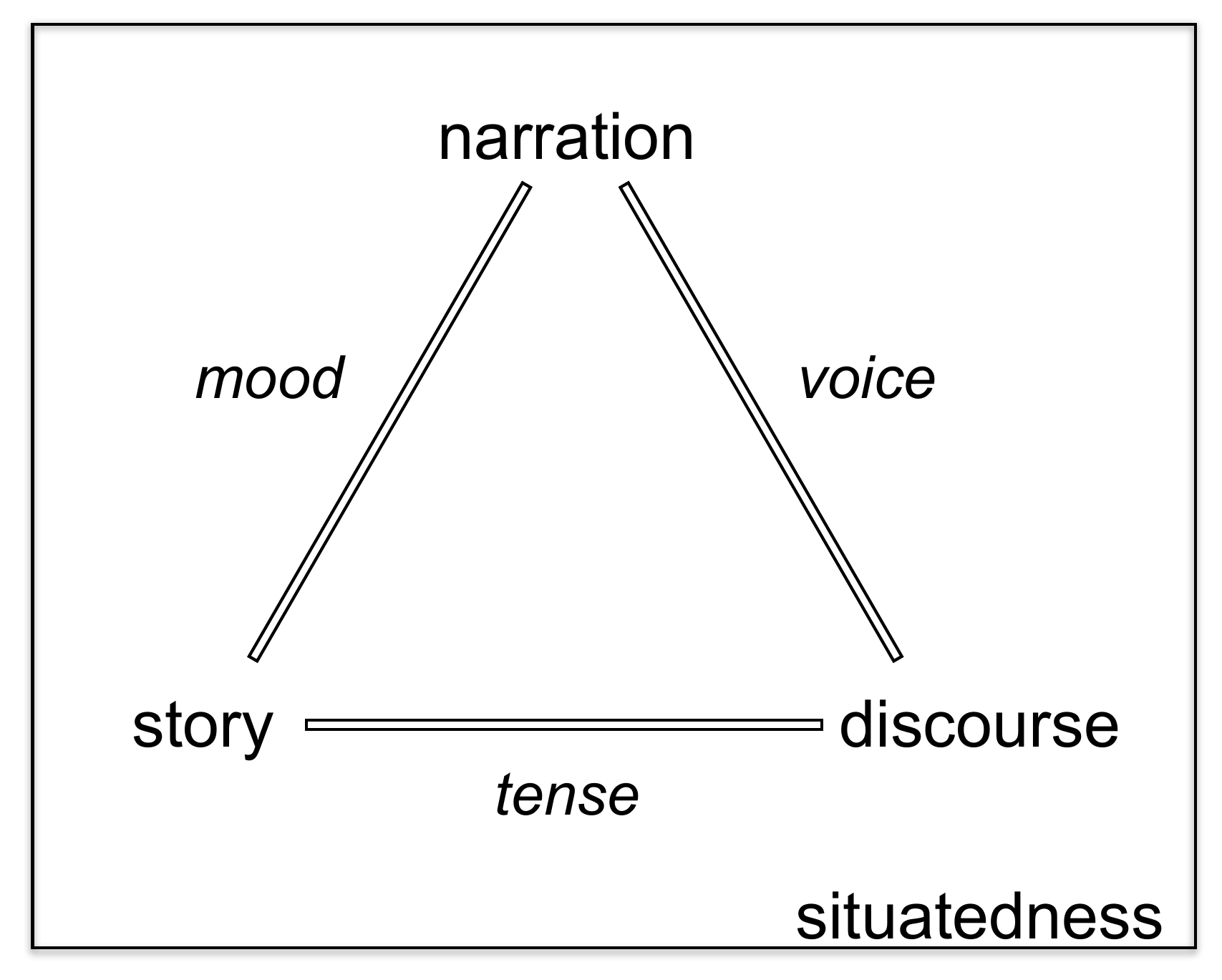}
\end{tcolorbox}
\caption{\textsc{NarraBench}'s primary theoretical foundations from \citet{genetteNarrativeDiscourseEssay1980} and \citet{herman2009basic}.}
\label{fig:triangle}
\end{figure}

To integrate these elements into a broader framework, we combine \citeauthor{genetteNarrativeDiscourseEssay1980}'s narrative triangle with \citeauthor{herman2009basic}'s notion of \emph{situatedness} (\autoref{fig:triangle}).
The Russian formalists \cite{tomashevsky1965russian} were the first to distinguish between \emph{story} (``what happened'') and \emph{discourse} (``how it was told''), to which \citet{genetteNarrativeDiscourseEssay1980} added a third dimension called \emph{narration}, which refers to the perspectival dimensions of narrative communication produced by the narrator's voice (``who speaks'').

Genette then introduced three further terms to capture the relationship between these dimensions, \emph{tense}, \emph{mood}, and \emph{voice}.
Genette extrapolates from the linguistic meanings of these terms to capture specific narratological features. 
These include aspects of time and the ordering of events (\emph{tense}); the relationship between eventfulness and description (\emph{mood}); and aspects related to perspective, such as point of view, dialogue, and focalization (\emph{voice}). 
\citet{herman2009basic} importantly adds a fourth dimension, \emph{situatedness}, which captures the social dimensions of narrative (medium, social context, interactive dimensions). 
We combine these elementary and higher-level dimensions to structure our taxonomy described in the next section.

\section{The \textsc{NarraBench} Taxonomy}
\label{sec:taxonomy}
\begin{table}[t!]
\centering
\footnotesize
\begin{tabular}{@{}lllc@{}}
\textbf{Dimension} & \textbf{Feature} & \textbf{Aspect} & \textbf{SMV} \\
\midrule
\multirow{20}{*}{\textbf{Story}}
& \multirow{8}{*}{Agent} & name & \textcolor{blue}{L}\kern2pt\textcolor{orange}{D}\kern2pt\textcolor{red}{D} \\
& & & \textcolor{blue}{G}\kern2pt\textcolor{orange}{H}\kern2pt\textcolor{red}{C} \\
& & role & \textcolor{blue}{G}\kern2pt\textcolor{orange}{H}\kern2pt\textcolor{red}{P} \\
& & attributes & \textcolor{blue}{L}\kern2pt\textcolor{orange}{D}\kern2pt\textcolor{red}{D} \\
& & & \textcolor{blue}{G}\kern2pt\textcolor{orange}{H}\kern2pt\textcolor{red}{C} \\
& & emotions & \textcolor{blue}{L}\kern2pt\textcolor{orange}{D}\kern2pt\textcolor{red}{P} \\
& & & \textcolor{blue}{G}\kern2pt\textcolor{orange}{H}\kern2pt\textcolor{red}{P} \\
& & motivation & \textcolor{blue}{L}\kern2pt\textcolor{orange}{D}\kern2pt\textcolor{red}{P} \\
& & & \textcolor{blue}{G}\kern2pt\textcolor{orange}{P}\kern2pt\textcolor{red}{P} \\
\cmidrule(lr){2-4}
& \multirow{3}{*}{Social Net} & interaction & \textcolor{blue}{L}\kern2pt\textcolor{orange}{D}\kern2pt\textcolor{red}{D} \\
& & connections & \textcolor{blue}{G}\kern2pt\textcolor{orange}{H}\kern2pt\textcolor{red}{D} \\
& & relationship & \textcolor{blue}{G}\kern2pt\textcolor{orange}{H}\kern2pt\textcolor{red}{C} \\
\cmidrule(lr){2-4}
& \multirow{4}{*}{Event} & event & \textcolor{blue}{L}\kern2pt\textcolor{orange}{D}\kern2pt\textcolor{red}{D} \\
& & & \textcolor{blue}{G}\kern2pt\textcolor{orange}{D}\kern2pt\textcolor{red}{C} \\
& & schema & \textcolor{blue}{G}\kern2pt\textcolor{orange}{H}\kern2pt\textcolor{red}{C} \\
& & causality & \textcolor{blue}{G}\kern2pt\textcolor{orange}{P}\kern2pt\textcolor{red}{P} \\
\cmidrule(lr){2-4}
& \multirow{7}{*}{Plot} & topic & \textcolor{blue}{G}\kern2pt\textcolor{orange}{H}\kern2pt\textcolor{red}{C} \\
& & plot & \textcolor{blue}{G}\kern2pt\textcolor{orange}{H}\kern2pt\textcolor{red}{P} \\
& & plotline & \textcolor{blue}{G}\kern2pt\textcolor{orange}{H}\kern2pt\textcolor{red}{C} \\
& & moral & \textcolor{blue}{G}\kern2pt\textcolor{orange}{H}\kern2pt\textcolor{red}{P} \\
& & obstacle & \textcolor{blue}{G}\kern2pt\textcolor{orange}{H}\kern2pt\textcolor{red}{P} \\
& & conflict & \textcolor{blue}{G}\kern2pt\textcolor{orange}{H}\kern2pt\textcolor{red}{P} \\
& & archetype & \textcolor{blue}{G}\kern2pt\textcolor{orange}{H}\kern2pt\textcolor{red}{C} \\
\cmidrule(lr){2-4}
& structure & plot arc & \textcolor{blue}{G}\kern2pt\textcolor{orange}{P}\kern2pt\textcolor{red}{C} \\
\cmidrule(lr){2-4}
& \multirow{4}{*}{Setting} & setting & \textcolor{blue}{L}\kern2pt\textcolor{orange}{D}\kern2pt\textcolor{red}{D} \\
& & & \textcolor{blue}{G}\kern2pt\textcolor{orange}{H}\kern2pt\textcolor{red}{C} \\
& & location & \textcolor{blue}{L}\kern2pt\textcolor{orange}{D}\kern2pt\textcolor{red}{D} \\
& & & \textcolor{blue}{G}\kern2pt\textcolor{orange}{D}\kern2pt\textcolor{red}{D} \\
\midrule
\multirow{8}{*}{\textbf{Discourse}}
& \multirow{5}{*}{Time} & duration & \textcolor{blue}{L}\kern2pt\textcolor{orange}{D}\kern2pt\textcolor{red}{D} \\
& & & \textcolor{blue}{G}\kern2pt\textcolor{orange}{P}\kern2pt\textcolor{red}{D} \\
& & & \textcolor{blue}{G}\kern2pt\textcolor{orange}{H}\kern2pt\textcolor{red}{D} \\
& & order & \textcolor{blue}{G}\kern2pt\textcolor{orange}{P}\kern2pt\textcolor{red}{D} \\
& & & \textcolor{blue}{G}\kern2pt\textcolor{orange}{H}\kern2pt\textcolor{red}{D} \\
\cmidrule(lr){2-4}
& \multirow{3}{*}{Revelation} & suspense & \textcolor{blue}{G}\kern2pt\textcolor{orange}{P}\kern2pt\textcolor{red}{P} \\
& & curiosity & \textcolor{blue}{G}\kern2pt\textcolor{orange}{P}\kern2pt\textcolor{red}{P} \\
& & surprise & \textcolor{blue}{G}\kern2pt\textcolor{orange}{P}\kern2pt\textcolor{red}{P} \\
\midrule
\multirow{8}{*}{\textbf{Narration}}
& \multirow{3}{*}{Perspective} & point of view & \textcolor{blue}{G}\kern2pt\textcolor{orange}{D}\kern2pt\textcolor{red}{D} \\
& & focalization & \textcolor{blue}{L}\kern2pt\textcolor{orange}{D}\kern2pt\textcolor{red}{D} \\
& & dialogue & \textcolor{blue}{L}\kern2pt\textcolor{orange}{D}\kern2pt\textcolor{red}{D} \\
\cmidrule(lr){2-4}
& \multirow{5}{*}{Style} & allusion & \textcolor{blue}{L}\kern2pt\textcolor{orange}{D}\kern2pt\textcolor{red}{P} \\
& & figurative & \textcolor{blue}{L}\kern2pt\textcolor{orange}{D}\kern2pt\textcolor{red}{P} \\
& & imageability & \textcolor{blue}{L}\kern2pt\textcolor{orange}{H}\kern2pt\textcolor{red}{P} \\
& & complexity & \textcolor{blue}{L}\kern2pt\textcolor{orange}{H}\kern2pt\textcolor{red}{P} \\
& & evaluative & \textcolor{blue}{L}\kern2pt\textcolor{orange}{D}\kern2pt\textcolor{red}{P} \\
\midrule
\multirow{6}{*}{\textbf{Situatedness}}
& \multirow{5}{*}{Paratext} & genre & \textcolor{blue}{G}\kern2pt\textcolor{orange}{H}\kern2pt\textcolor{red}{C} \\
& & author & \textcolor{blue}{G}\kern2pt\textcolor{orange}{D}\kern2pt\textcolor{red}{D} \\
& & date & \textcolor{blue}{G}\kern2pt\textcolor{orange}{D}\kern2pt\textcolor{red}{D} \\
& & medium & \textcolor{blue}{G}\kern2pt\textcolor{orange}{D}\kern2pt\textcolor{red}{D} \\
& & platform & \textcolor{blue}{G}\kern2pt\textcolor{orange}{D}\kern2pt\textcolor{red}{D} \\
\cmidrule(lr){2-4}
& Motivation & intent & \textcolor{blue}{G}\kern2pt\textcolor{orange}{H}\kern2pt\textcolor{red}{P} \\
\end{tabular}
\caption{The complete \textsc{NarraBench} taxonomy. We indicate \textbf{SMV} (\textcolor{blue}{S}cale-\textcolor{orange}{M}ode-\textcolor{red}{V}ariance) with: \textcolor{blue}{L/G}=local/global, \textcolor{orange}{D/H/P}=discrete/holistic/progressive, \textcolor{red}{D/C/P}=deterministic/consensus/perspectival.}
\label{tab:mini}
\end{table}

Our taxonomy focuses on two core levels of narrative understanding. The first is a hierarchically arranged set of fifty \textbf{narrative aspects} that can be mapped to the twelve \emph{narrative features} shown in \autoref{fig:benchmark-distributions}. We group these aspects under the four principal \emph{narrative dimensions} depicted in \autoref{fig:triangle}. These dimensions form the root of our taxonomy.

The second level of our taxonomy focuses on \textbf{evaluation criteria}. These include textual \emph{scale} (local, global, meso), \emph{mode} (discrete, progressive, holistic judgments), and the expected \emph{variance} of potential answers (deterministic, consensus, perspectival). We present the taxonomy in its entirety in \autoref{tab:taxonomy}, and a condensed version in \autoref{tab:mini}.

\subsection{Narrative Aspects}

\subsubsection{STORY}
Story captures all aspects of a narrative that relate to storyworld content. We pay attention to:

\paragraph{Agents.} Narratives foreground the lived experiences of agentic entities (i.e., characters) \cite{fludernik2002towards}. Agent detection and inventorying \emph{character names} are an essential first step. Higher-level synthesis of characters around their narrative \emph{roles} and \emph{attributes} is necessary for understanding the functional position of characters within narratives \cite{proppMorphologyFolktaleSecond1968}. We also include theory-of-mind related tasks such as \emph{narrative emotion} labelling and character \emph{motivation} understanding. Interpreting agent intentionality and inner experience is a central component of narrative understanding \cite{zunshine2006we, mar2011neural, oatley2016fiction}. 

\paragraph{Social Networks.} Agents in a narrative exist in a particular social configuration.
Here we divide social networks into three core aspects: \emph{interactions} (local social events), \emph{connections} (who does each character know), and \emph{relationships} (the social types of connections, including employment, family, civil, romantic, and friendship-based connections) \cite{labatut2019extraction}.

\paragraph{Events.} Narratives are structured around agents who experience change over time. Such change arises through \emph{action}, where an action and its corresponding effect together form a \emph{causal frame}. While narratologists debate the precise boundaries of a frame within a story, it is broadly accepted that reconstructing the underlying chain of actions and their causal relations is central to recovering the narrative’s event structure or \emph{schemas} \cite{schankScriptsPlansGoals1975, chambersUnsupervisedLearningNarrative2009}.

\paragraph{Plot.} According to \citet{Kukkonen2014Plot}, ``The term \emph{plot} designates the ways in which the events and characters’ actions in a story are arranged and how this arrangement in turn facilitates identification of their motivations and consequences.'' In a technical sense, plot sits between story and discourse, but for our purposes we consider it as part of the story dimension. Plot can thus be subdivided into \emph{topics} (core thematic concerns), \emph{plot lines} (sub-plots), key structural forces like narrative \emph{obstacles} or \emph{conflicts}, central value lessons (i.e., \emph{morals}), as well as larger schematic structures like \emph{archetypes}. 

\paragraph{Structure.} \citet{Kukkonen2014Plot} makes a core distinction between plot as a function of ``global structure'' (our primary definition above) and ``progressive structuration,'' that is, aspects of plot that depend on the sequencing of events for their meaning. We reserve these progressive aspects for the \emph{discourse} dimension, with the exception of \emph{plot arcs}, which capture structural elements related to events such as fortune \cite{reaganEmotionalArcsStories2016, elkinsShapesStoriesSentiment2022, mcadams2001bad}, reversal \cite{knight2024narrative}, and denouement \cite{freytag1895technique}. Understanding the reliance or inability of LLMs to produce certain plot arc structures is important for shaping reader expectations.

\paragraph{Setting.} The spatial distribution of perception and movement within narratives is a key aspect of defining audiences' cognitive orientation.  Where narrative theory has introduced concepts such as the  ``chronotope'' \cite{bakhtin2010dialogic} or ``possible worlds'' \cite{ryan2016narrating}, we use the more integrative term of \emph{setting} to capture both the holistic understanding of narrative space and the particular location of a given narrative moment.

\subsubsection{NARRATION}

Narratives require narrators. According to \citet{genetteNarrativeDiscourseEssay1980}, ``who speaks'' plays a central role in governing the selection and communication of narrative content, which will in turn impact how audiences interpret the narrative.

\paragraph{Perspective.} Narrative perspective encompasses classical notions of \emph{point-of-view} (technically referred to as homo- and heterodiegetic narrators, 1P/3P respectively) as well as the subtler category of \emph{focalization}, a local sense of the eyes or mind through which certain events are perceived. We also include \emph{dialogue} detection and speaker attribution here to capture aspects of direct speech that are central to narratives.

\paragraph{Style.} We use style to capture rhetorical and intertextual aspects of narratives that bear on their meaning. \emph{Allusion} refers to inter-textual references \cite{kristeva1980desire, genette1997palimpsests}, while \emph{figurative} language and \emph{imageability} are concerned with the symbolic and/or perceptual dimensions of narrative communication. We also include \emph{complexity} as a measure of reading level and social capital and \emph{evaluative} language as a way of capturing narrative meta-awareness \cite{labov1997narrative}. 

\subsubsection{DISCOURSE}

Where \emph{narration} captures aspects of perspective and style, \emph{discourse} reflects organizational choices made in \emph{how} events are rendered.

\paragraph{Time.} Narratives take place over time. But time is not necessarily represented in a chronological manner, nor is time represented in an equitable manner. Narrators may decide to dedicate greater fidelity to certain events, or reflect back on prior events or foreshadow future events. Following \citet{genetteNarrativeDiscourseEssay1980}, we focus on both \emph{duration} and \emph{order} with respect to time.

\paragraph{Revelation.} How information is revealed to the reader over the course of narrative time fundamentally shapes readers' affective state \cite{brewer1982stories}. Here we focus on the classic tripartite scheme of reader \emph{suspense}, \emph{curiosity}, and \emph{surprise} to capture the assessment of narrative revelation over time.

\subsubsection{SITUATEDNESS}

A narrative is always communicated in a real context. This social context inflects both authorial choice and audience expectations.

\paragraph{Paratext.} What theorists call the \emph{paratext} are those elements of a narrative that reflect its real-world context. This includes aspects like \emph{genre}, \emph{author}, \emph{date}, and \emph{medium}. These dimensions help inform reader interpretations of the material.

\paragraph{Motivation.} The ``why'' behind the telling of a narrative is captured by the author's intent. Narratives may be told for a variety of reasons, but being able to reasonably infer authorial \emph{intent} is an essential task of narrative understanding.

\subsection{Evaluation Criteria}

We divide our evaluation framework into three primary dimensions.

\paragraph{Scale.} When assessing narrative understanding, it is important to identify the textual \emph{scale} of a particular judgment. Some features, such as character interactions, are always \emph{local} in nature, because they are located at a discrete point in the narrative. Others are \emph{global}, such as character roles or plot summaries. These require abstractive reasoning from a narrative unit (the entire narrative or a meaningful sub-portion). We reserve the term \emph{meso} for features that depend on reasoning around temporal sequences, such as duration or suspense. 

\paragraph{Mode.} Answers might take various forms (e.g., a string, an integer, or a boolean) depending on the measurement. We call this the \emph{mode}. The mode of the answer refers to whether the value itself is \emph{discrete} in nature (meaning it is a single, token-based value), \emph{progressive} (a collection of values forming a time series), or \emph{holistic} (a comprehensive answer in the form of a string written in natural language). The mode plays a part in determining what form of automatic evaluation is admissible.

\paragraph{Variance.} This dimension captures assumptions about  interpretive freedom. Some features may require single, correct answers, such as a character name. We label these \emph{deterministic}. Other features are more open-ended, such as narrative emotion or narrative intent. We label these features \emph{perspectival} when there is a distribution of possible answers that is also contextually dependent. We finally use \emph{consensus} where there are clear central tendencies in a distribution of answers. 

\subsection{Taxonomy Construction}

We identify fifty total permutations of evaluation criteria and narrative attributes that are well-founded in the theoretical literature and critical for narrative understanding tasks (\autoref{tab:taxonomy}). As we highlight in our conclusion, a core aspect of this framework is its expandability. We capture the foundation of each permutation with a single question, of possible use for model testing.

\section{Benchmark Survey}
\label{sec:survey}
\begin{table}[t]
\centering
\small
\begin{tabular}{lrr}
  \textbf{Quality} & \textbf{Distance} & \textbf{Benchmark Count} \\
  \midrule
  Good & 0 & 10 \\
  Decent  & 1 & 14 \\
  Poor & 2 & 10 \\
  Bad {\small (ignored)} & 3 & 5 \\
  Missing Data & - & 39 \\
  \bottomrule
  \emph{Total} & & 78
\end{tabular}
\caption{Overall benchmark alignment distribution. Note we consider only those benchmarks with an edit distance of 2 or below.}
\label{tab:distance-dist}
\end{table}

The purpose of the survey is to adjudicate to what degree currently available LLM benchmarks satisfy the aspects of narrativity identified above.
For our purposes, we define a benchmark to be an annotated dataset or testing technique made available for testing language model accuracy on some qualitative task. The benchmark should:

\begin{itemize}
    \item Yield results useful for adjudicating model performance of one or more of our twelve \emph{narrative features}.
    \item Have an existing repository available for providing code and data.
    \item Provide convenience functions (e.g., a script or software framework) for testing on arbitrary language models using a standard API.
    \item Serve as a metric for comparing the relative performance of different models. It should not expect a particular model family.
\end{itemize}

\begin{figure}
\begin{tcolorbox}[colback=white,colframe=black,boxrule=0.7pt,left=4pt,sharp corners]
\sffamily\footnotesize\centering
\begin{tabular}{@{}l l@{\hspace{6pt}} l@{}}
\multirow{9}{*}[38pt]{\textbf{Agent}} 
& role & \href{https://github.com/KaiHe-better/Crab}{Crab}, \href{https://github.com/OFA-Sys/Ditto}{Ditto} \\
& attributes & \href{https://github.com/Wellesley-EASEL-lab/AustenAlike}{AustenAlike}, \href{https://github.com/adobe-research/NoLiMa}{NoLiMa} \\
& emotions & \href{https://github.com/llm-for-emotion/culemo}{CULEMO}, \href{https://github.com/EQ-bench/EQ-Bench}{EQ-Bench} \\
& & \href{https://github.com/zhchen18/ToMBench}{ToMBench} \\
& & \href{https://github.com/CUHK-ARISE/EmotionBench}{EmotionBench} \\
& motivation & \href{https://github.com/seacowx/OpenToM}{OpenToM} \\
& & \href{https://github.com/skywalker023/fantom}{FANTOM} \\ 
& & \href{https://github.com/microsoft/AnthropomorphicIntelligence/blob/main/MotiveBench/README.md}{MOTIVEBENCH} \\ 
& & \href{https://github.com/thu-coai/CharacterBench}{CHARACTERBENCH} \\[3pt]
\rowcolor{gray!15} \multirow{2}{*}[4pt]{\textbf{Social Net}} 
& connections & \href{https://github.com/kilian-group/phantom-wiki}{PhantomWiki} \\
\rowcolor{gray!15} & relationship & \href{https://github.com/kwai/DialogBench}{DialogBench} \\[3pt]
\textbf{Plot}
& topic & \href{https://github.com/ServiceNow/repliqa}{RepLiQA} \\
& plot & \href{https://github.com/kabirahuja2431/FlawedFictions}{FlawedFictions} \\
& plotline & \href{https://github.com/melaniesubbiah/storysumm}{StorySumm} \\
& moral & \href{https://github.com/agiresearch/MoralBench}{MoralBench}\\
& & \href{https://huggingface.co/datasets/cardiffnlp/Morables}{MORABLES} \\[3pt]
\rowcolor{gray!15} \textbf{Setting} & location & \href{https://zenodo.org/records/16670471}{Dataset-GSS}  \\
 \textbf{Revelation}
& suspense & \href{https://github.com/zhaochen0110/conflictbank}{ConflictBank} \\ \rowcolor{gray!15}
\multirow{2}{*}[4pt]{\textbf{Perspective}} 
& dialogue & \href{https://github.com/johndmendonca/soda_eval?tab=readme-ov-file}{SODA-EVAL}, \href{https://github.com/Neph0s/CoSER}{CoSER} \\
\multirow{4}{*}[16pt]{\textbf{Time}} 
& order & \href{https://gitlab.ub.uni-bielefeld.de/s.kenneweg/TRaVelER}{TRaVelER}, \href{https://huggingface.co/datasets/baharef/ToT}{ToT}, \\
& & \href{https://github.com/THU-KEG/MAVEN-ERE}{MAVEN-ERE}, \href{https://github.com/EternityYW/TRAM-Benchmark}{TRAM} \\[2pt]
\rowcolor{gray!15} \textbf{Motivation}
& intent & \href{https://huggingface.co/datasets/APauli/Persuasive-Pairs}{Persuasive Pairs} \\
\multirow{3}{*}[8pt]{\textbf{Paratext}} 
& genre & \href{https://github.com/uchidalab/book-dataset}{BookCover30} \\
& author & \href{https://github.com/bit-ml/VeriDark}{VeriDark}, \href{https://github.com/YashSaxena21/REASONS}{REASONS} \\
& date & \href{https://huggingface.co/datasets/hereldav/TimeAware}{TimeAware} \\[2pt]
\end{tabular}
\end{tcolorbox}
\caption{Aligned open benchmarks for narrative understanding tasks identified in this survey.}
\label{tab:bench}
\end{figure}

Following these guidelines, we surface 78 potential benchmarks published over the last twelve years via relevant keyword searches on Google Scholar.
Filtering for open data reduces this count to 39 benchmarks.
We then adjudicate how well each benchmark matches the respective evaluation attributes for each hierarchal attribute triple (e.g. local-discrete-deterministic), measuring the edit distance.
From this value we assign alignment ratings with our taxonomy of ``good,'' (a distance of 0), ``decent,'' (1), ``poor,'' (2), and finally  ``bad'' (3), resulting in the distribution seen in \autoref{tab:distance-dist}.
Retaining all benchmarks with a quality rating of ``poor'' and above leaves us with 34 benchmarks which we present in \autoref{tab:bench} and with a complete rating breakdown in \autoref{tab:survey}.
We now highlight trends observed when conducting our survey.

\paragraph{Increasing popularity.}
Of the 39 benchmarks we surveyed, half were published in 2024 or 2025, with the majority in 2025.
As LLM use increases, so too do benchmarking efforts in an attempt to provide metrics for questioning whether advancements in ML techniques are translating to performance increases in downstream capabilities.
This increasing popularity carries with it the risk of increasing the signal-to-noise ratio in the benchmarking space.
This risk highlights the need to build unified theoretical frameworks to guide these efforts.

\paragraph{Moving beyond classification tasks.} Of the 39 benchmarks surveyed, 27 rely on classification or multi-label prediction, while 12 evaluate open-ended text generation. 
Causal LLMs are generative models. Evaluating their generated text therefore offers a direct window into their underlying biases and proficiencies, as has been increasingly adopted  \citep{lucyGenderRepresentationBias2021,hickeZeroBodyProblem2025}. 
Expanding such generative benchmarking approaches would enable the study of model responses in more holistic and perspectival ways.

\begin{table*}[t!]
\centering
\small
\begin{tabular}{@{}lrrrrrrr|r@{}}
 & \textbf{Attributes} & \textbf{Emotional} & \textbf{Connections} & \textbf{Plotline} & \textbf{Order} & \textbf{Order} & \textbf{Order} & \\
\textbf{Model}  & \textit{austenalike} & \textit{culemo} & \textit{phantomwiki} & \textit{storysumm} & \textit{tot} & \textit{tram} & \textit{traveler} & \textbf{Mean} \\
\midrule
Gemma-3 12B & \textbf{3.2} & 41.0 & 17.8 & 40.6 & 3.8 & 67.0 & 54.0 & 32.5 \\
GPT-OSS 20B & 0.0 & \textbf{100.0} & 2.6 & \textbf{62.5} & 1.2 & 0.0 & 19.0 & 26.5 \\
Llama-3 8B & 2.1 & 28.5 & \textbf{23.0} & 55.2 & \textbf{31.8} & \textbf{59.5} & \textbf{61.0} & \textbf{37.3} \\
\end{tabular}
\caption{Results across all well-conforming benchmarks and models. Accuracy is reported as percentages.}
\label{tab:results}
\end{table*}

\paragraph{An overemphasis on story.}
We identified 19 benchmarks that address our story-specific features.
Compared to benchmarks identified for narration (2), discourse (5), and situatedness (5), this suggests current benchmarking efforts overemphasize the story as a target. One of the principal aims of our theoretical framework is to foreground the complexity of narrative communication and the importance of narrative perspective, discourse structure, and social context for surfacing narrative meaning.

\paragraph{Lack of event-specific benchmarking.} While we identified many existing annotated datasets capturing event-related tasks \citep{bakerBerkeleyFrameNetProject1998,simsLiteraryEventDetection2019,tangDiscourseNarrativeKnowledge2021,wangUncoveringSurprisingEvent2022}, none were expressly equipped for benchmarking as outlined in our requirements.
These annotated datasets, while valuable, still need to be converted to the criteria outlined in the beginning of \autoref{sec:survey} to be considered benchmarks.

\paragraph{Lack of style-specific benchmarking.}
We did not identify any benchmarks for allusion detection, figurative language production, imageability, complexity, or evaluative language use in language models. Figurative use of language is an important dimension for understanding narrative meaning. Benchmarking how effectively large language models recognize this intentional use of language is an important aspect of understanding their general narratological abilities.
We encourage researchers to draw on existing methodologies and datasets in the fields of stylometry and metaphor detection when designing benchmarks for these aspects.

\paragraph{Lack of subjectivity in responses.}
All but two of the benchmarks constructed their ground truth in a deterministic manner, meaning LLMs \emph{must} must correctly select an answer from a set of possible responses.
As we note in \autoref{sec:relatedwork}, benchmark developers are now phasing this mode of judgment out in favour of allowing for a multiplicity of responses. 
This method is more accurate when considering ground truth is often based on annotations from multiple annotators, meaning ground truth begins as a distribution of responses.
Non-deterministic responses likewise more proficiently capture the subjective quality of audience response, in that no two audience members will respond to a narrative in precisely the same way.
The same expectation must be afforded to LLMs if their abilities are to be fairly judged \citep{plank-2022-problem, pichler-etal-2025-evaluating}.

\paragraph{Per-token responses are still needed.}
Early statistical NLP techniques were often built on lexical corpora attributing per-token scores for various linguistic phenomena.
This enabled algorithms for assessing the phenomena over particular substrings, important for researchers interested in identifying unexpected failure modes over text input \citet{linROUGEPackageAutomatic2004}.
The convenience of passing in an entire text to a LLM and receiving back whole answers has motivated developers to prefer holistic/global assessments over discrete responses.
This has the benefit of allowing language models the space to respond as they naturally would to a query of a similar nature, but evaluating per-token responses can elicit a more nuanced understanding of how language models represent complex semantics.
We argue both approaches are important for narrative understanding and encourage benchmark developers to continue to value token-level benchmarking.

\paragraph{Global languages are not well attended.}
Four of the 39 benchmarks we shortlisted were multilingual, with the rest implementing their test in English.
This gap leaves open the question of whether language models can properly represent narrative dimensions in non-English languages (and especially low-resource languages).
We encourage benchmark developers to seek platforms that provide human assessments from global perspectives alongside multilingual narrative resources.

\paragraph{Lack of open data for reproducibility.}
We finally note only 39 of the original 78 narrative-aligned benchmarks we identified made their code and data available in a freely-accessible repository.
While certain papers made their rationale known (e.g. their benchmark depends on in-copyright text), other papers offered links leading to dead webpages. Adhering to best practices in data management, i.e. the use of long-term repositories with stable DOIs, is an essential dimension for future benchmarking initiatives.

\paragraph{Advancing benchmarks towards multimodal inputs.}
Narratives exist in multiple modalities.
Narratological research in recent decades has developed theoretical frameworks for describing how narratives behave in music, imagery, paintings, and movies \citep{fludernikPostclassicalNarratologyApproaches2010}.
Approximately 5\% of our surveyed benchmarks were either multimodal or non-textual in their inputs, testing vision language models (VLMs) on their visual understanding abilities.
Future benchmarks will want to combine these approaches to work towards a framework for describing multimodal narrative concepts.

\section{Testing Harness}
\textsc{NarraBench} is a novel attempt at reconfiguring existing benchmarks to fulfill theoretical goals, and as such can be implemented as an expandable collection of relevant benchmarks. To guide future implementations we have constructed a testing harness containing the most suitable benchmarks identified in our study, showcasing sample scores from a series of popular open-weight LLMs in \autoref{tab:results}.\footnote{We make our harness available \href{https://github.com/srhm-ca/NarraBench}{here.}}
Our results suggest non-uniform performance across models and specific narrative aspects, a result not immediately obvious when making use of a specific benchmark as an indicator for story understanding.
We encourage all interested to identify or develop further benchmarks in areas not yet covered as per our initial survey.
Distributing a stable implementation is one important way \textsc{NarraBench} will remain an evolving and relevant resource moving into the future.

\section{Conclusion}
To help guide and centralize growing efforts in narrative benchmarking, we present \textsc{NarraBench}, the first comprehensive narratological taxonomy for benchmark developers to integrate efforts around a shared, holistic understanding of narrative communication.
\textsc{NarraBench} distills decades of literary theory down to four fundamental narrative dimensions: \emph{story}, \emph{narration}, \emph{discourse} (or the structure of the narrative), and \emph{situatedness} (the context in which the narrative exists). We further divide these dimensions into twelve primary \emph{features} and fifty overall \emph{aspects}, described in \autoref{tab:taxonomy}.

As a first step towards constructing our envisioned framework, we conducted a first-ever survey of existing narrative benchmarks to identify how contemporary benchmarking efforts align with the \textsc{NarraBench} taxonomy (\autoref{tab:survey}). 
We identify 39 open-data benchmarks whose qualitative targets roughly match one of the fifty permutations defined in our taxonomy, suggesting  existing benchmarks satisfy $\approx$27\% of the \textsc{NarraBench} taxonomy as indicated in \autoref{fig:benchmark-distributions}.
This gap suggests opportunities for researchers to produce new tasks.

Central to our effort is the notion of an \emph{expandable} framework for narrative benchmarking. Our high-level Big-4 dimensions allow for growth at all three subsidiary levels. Like other large benchmarking efforts, \textsc{NarraBench} can be seen as a benchmark of benchmarks organized around a core theoretical foundation. We encourage researchers in the field to both fill in the missing holes identified by our survey (as shown in \autoref{tab:survey}) and also propose novel features and assessments. Providing one theoretical description of narrative understanding in the form of \textsc{NarraBench} offers benchmark developers the opportunity to consult a central resource to ensure efforts do not replicate prior work and address core methodological shortcomings.

\paragraph{Next steps.}
Researchers are increasingly focusing on the value of narrative understanding and narrative generation.
Our testing harness will allow community members to consult the current state of the art, and contribute where possible.
Now is an optimal time to begin the work of consolidation to provide LLM researchers with clear assessment criteria around one of the most fundamental ways humans relate to each other: storytelling. 

\section*{Limitations}

While \textsc{NarraBench} provides the first comprehensive taxonomy for narrative benchmarking, several limitations should be acknowledged.
First, the framework is grounded in a particular lineage of narrative theory—what is known as the classical model—which privileges compositional and communicative dimensions of narrative (story, narration, discourse, situatedness). This excludes alternative conceptions of narrativity, such as those derived from cognitive narratology, rhetorical theory, or postclassical approaches emphasizing affect, ideology, or embodiment. As a result, \textsc{NarraBench} captures a specific, though widely applicable, sense of computational narrative understanding.

Second, the benchmark survey is constrained by the availability and accessibility of existing resources. Of the 78 benchmarks identified, only half provided open data or usable repositories. This reliance on public datasets may bias coverage toward English-language and Western-centric corpora, underrepresenting non-Western narrative traditions and low-resource languages. Likewise, many of the included benchmarks are text-based, limiting our ability to generalize conclusions about multimodal or cross-media narratives, which are increasingly central to human storytelling practices.

Third, \textsc{NarraBench} currently formalizes narrative understanding in a way that presupposes task modularity—that discrete components such as perspective, style, or revelation can be isolated and measured independently. While this approach enables systematic comparison, it underplays the interdependence and nonlinearity of narrative meaning-making. Future iterations of the framework could incorporate compositional or causal dependencies across dimensions, testing how models integrate multiple narrative features jointly rather than in isolation.

Fourth, because the taxonomy is conceptual rather than empirical, its coverage and weighting of dimensions remain interpretive. Our survey indicates that existing benchmarks cover roughly 27\% of the proposed taxonomy, but the relative importance of each category for narrative comprehension has not yet been empirically validated. The framework should therefore be viewed as a scaffolding for community refinement rather than a definitive account of narrative intelligence.

Finally, despite its explicit focus on perspectival and consensus-based evaluation, \textsc{NarraBench} inherits many of the broader evaluation challenges facing LLM benchmarking: the instability of LLM-as-a-judge methods, limited reproducibility of generative scoring, and uncertainty about how human interpretive diversity should be represented in gold standards. Expanding community participation, diversifying annotation pipelines, and incorporating non-deterministic scoring schemes will be essential to properly realizing narrative benchmarking.

In sum, \textsc{NarraBench} represents a first synthesis—a structured, extensible foundation for assessing narrative understanding—while acknowledging that both its theoretical premises and empirical coverage are partial. Its value lies in providing a coherent starting point for an evolving research community to debate, test, and revise what narrative understanding should mean for NLP and ML.

\section*{Ethical Considerations}

Narrative is among the most powerful instruments of human communication. It can promote empathy, understanding, and collective meaning-making, but it can equally serve as a vehicle for misinformation, propaganda, and ideological manipulation. As large language models increasingly generate and interpret narratives at scale, the ethical implications of how these capacities are benchmarked become critical. A benchmark that rewards coherence or emotional resonance without attention to factuality, bias, or intent risks amplifying the very distortions that narrative can produce.

\textsc{NarraBench} is designed in part to mitigate these risks by making explicit the multidimensional nature of narrative understanding—its agents, events, perspectives, and social contexts. By foregrounding interpretive variance and perspectival alignment, the framework aims to encourage evaluations that account not only for textual quality but also for the moral and epistemic consequences of narrative production. Nevertheless, benchmarking itself is never neutral: the selection of narratives, the construction of evaluation criteria, and the choice of what counts as ``understanding'' all carry normative assumptions.

We therefore view narrative benchmarking as an ethical task of representation. Establishing shared standards for how models comprehend and reproduce stories is essential for preventing the misuse of narrative generation in contexts that shape public opinion, cultural memory, personal well-being, and political discourse. Getting narrative benchmarking right means ensuring that models are evaluated not only for their formal proficiency but also for their capacity to reflect the diversity, accountability, and responsibility inherent in human storytelling.

\bibliography{custom}

\appendix

\section{Complete Taxonomy}
We present the complete \textsc{NarraBench} taxonomy together with example questions in \autoref{tab:taxonomy}.

\begin{table*}[t]
\centering
\small
\begin{tabular}{@{}llllllp{5cm}@{}}
\toprule
\textbf{Dimension} & \textbf{Feature} & \textbf{Aspect} & \textbf{Scale} & \textbf{Mode} & \textbf{Variance} & \textbf{Question} \\
\midrule
\multirow{20}{*}{\textbf{Story}} 
& \multirow{8}{*}{Agent} & name & local & discrete & deterministic & \emph{Who are the characters in the text?} \\
& & & global & holistic & consensus & \emph{Who are the main characters in the text?} \\
& & role & global & holistic & perspectival & \emph{What is the character's role in the text?} \\
& & attributes & local & discrete & deterministic & \emph{What attributes does this character have?} \\
& & & global & holistic & consensus & \emph{What attributes does this character have?} \\
& & emotions & local & discrete & perspectival & \emph{What is the character feeling right now?} \\
& & & global & holistic & perspectival & \emph{What are the central emotional states?} \\
& & motivation & local & discrete & perspectival & \emph{Why is the character doing this right now?} \\
& & & global & progressive & perspectival & \emph{What motivates this character?} \\
\cmidrule(lr){2-7}
& \multirow{3}{*}{Social Net} & interaction & local & discrete & deterministic & \emph{How are these two characters interacting?} \\
& & connections & global & holistic & deterministic & \emph{Who does the character know?} \\
& & relationship & global & holistic & consensus & \emph{What is the relationship type?} \\
\cmidrule(lr){2-7}
& \multirow{4}{*}{Event} & event & local & discrete & deterministic & \emph{What is happening?} \\
& & & global & discrete & consensus & \emph{What happened?} \\
& & schema & global & holistic & consensus & \emph{What is the narrative schema?} \\
& & causality & global & progressive & perspectival & \emph{What caused this event?} \\
\cmidrule(lr){2-7}
& \multirow{7}{*}{Plot} & topic & global & holistic & consensus & \emph{What are the topics of this story?} \\
& & plot & global & holistic & perspectival & \emph{What is the plot summary?} \\
& & plotline & global & holistic & consensus & \emph{What happened in this plotline?} \\
& & moral & global & holistic & perspectival & \emph{What is the moral of the story?} \\
& & obstacle & global & holistic & perspectival & \emph{What is the central negative force?} \\
& & conflict & global & holistic & perspectival & \emph{What is the central conflict?} \\
& & archetype & global & holistic & consensus & \emph{What is the narrative archetype?} \\
\cmidrule(lr){2-7}
& structure & plot arc & global & progressive & consensus & \emph{What is the plot arc structure?} \\
\cmidrule(lr){2-7}
& \multirow{4}{*}{Setting} & setting & local & discrete & deterministic & \emph{What is the setting?} \\
& & & global & holistic & consensus & \emph{What is the setting?} \\
& & location & local & discrete & deterministic & \emph{Where is this taking place?} \\
& & & global & discrete & deterministic & \emph{What locations has the story visited?} \\
\midrule
\multirow{8}{*}{\textbf{Discourse}}
& \multirow{5}{*}{Time} & duration & local & discrete & deterministic & \emph{How much time is passing?} \\
& & & global & progressive & deterministic & \emph{How much time since the previous scene?} \\
& & & global & holistic & deterministic & \emph{How much time has passed?} \\
& & order & global & progressive & deterministic & \emph{Does this scene come before or after?} \\
& & & global & holistic & deterministic & \emph{Does this story tell events out of order?} \\
\cmidrule(lr){2-7}
& \multirow{3}{*}{Revelation} & suspense & global & progressive & perspectival & \emph{Is key information being withheld?} \\
& & curiosity & global & progressive & perspectival & \emph{Are causal antecedents being withheld?} \\
& & surprise & global & progressive & perspectival & \emph{Is key information suddenly revealed?} \\
\midrule
\multirow{8}{*}{\textbf{Narration}}
& \multirow{3}{*}{Perspective} & point of view & global & discrete & deterministic & \emph{Who is telling? (1P, 2P, 3P)} \\
& & focalization & local & discrete & deterministic & \emph{From whose POV are we seeing events?} \\
& & dialogue & local & discrete & deterministic & \emph{Who speaks? Identify speakers.} \\
\cmidrule(lr){2-7}
& \multirow{5}{*}{Style} & allusion & local & discrete & perspectival & \emph{What texts is this alluding to?} \\
& & figurative & local & discrete & perspectival & \emph{Is this using figurative language?} \\
& & imageability & local & holistic & perspectival & \emph{How well can you imagine this scene?} \\
& & complexity & local & holistic & perspectival & \emph{How complex is the sentence structure?} \\
& & evaluative & local & discrete & perspectival & \emph{Is this engaging in evaluative discourse?} \\
\midrule
\multirow{6}{*}{\textbf{Situatedness}}
& \multirow{5}{*}{Paratext} & genre & global & holistic & consensus & \emph{What is the genre?} \\
& & author & global & discrete & deterministic & \emph{Who is/are the author(s)?} \\
& & date & global & discrete & deterministic & \emph{What is the date of creation/publication?} \\
& & medium & global & discrete & deterministic & \emph{What medium?} \\
& & platform & global & discrete & deterministic & \emph{What platform?} \\
\cmidrule(lr){2-7}
& Motivation & intent & global & holistic & perspectival & \emph{What is the author's intent?} \\
\bottomrule
\end{tabular}
\caption{The complete \textsc{NarraBench} taxonomy of narrative understanding tasks, systematically derived from narratological theory. Each task is defined by six attributes: dimension, feature, aspect, scale, mode, and variance.}
\label{tab:taxonomy}
\end{table*}

\section{Survey Results}
We present our full survey results in \autoref{tab:survey}.

\begin{table*}[t]
\centering
\small
\setlength{\tabcolsep}{3pt}
\begin{tabular}{@{}lllrrrr@{}}
\toprule
\textbf{Feature} & \textbf{Aspect} & \textbf{Benchmark} & \textbf{Scale} & \textbf{Mode} & \textbf{Variance} & \textbf{Citation} \\
\midrule
\multirow{11}{*}{\textbf{Agent}} 
 & name & --- & --- & --- & --- & --- \\
& role & \href{https://github.com/KaiHe-better/Crab}{Crab} & \green{global} & \green{holistic} & \red{consensus} (P) & \citet{heCrabNovelConfigurable} \\
& & \href{https://github.com/OFA-Sys/Ditto}{Ditto} & \green{global} & \green{holistic} & \red{deterministic} (P) & \citet{luLargeLanguageModels2024} \\
& attributes & \href{https://github.com/Wellesley-EASEL-lab/AustenAlike}{AustenAlike} & \green{global} & \green{holistic} & \green{consensus} & \citet{yangEvaluatingComputationalRepresentations2024} \\
& & \href{https://github.com/adobe-research/NoLiMa}{NoLiMa} & \green{global} & \green{holistic} & \red{deterministic} (C) & \citet{modarressiNoLiMaLongContextEvaluation2025} \\
& emotions & \href{https://github.com/llm-for-emotion/culemo}{CULEMO} & \green{global} & \green{holistic} & \green{perspectival} & \citet{belayCULEMOCulturalLenses2025} \\
& & \href{https://github.com/EQ-bench/EQ-Bench}{EQ-Bench} & \green{global} & \green{holistic} & \red{deterministic} (P) & \citet{paechEQBenchEmotionalIntelligence2024} \\
& & \href{https://github.com/zhchen18/ToMBench}{ToMBench} & \green{global} & \green{holistic} & \red{deterministic} (P) & \citet{chenMBENCHBenchmarkingTheory} \\
& & \href{https://github.com/CUHK-ARISE/EmotionBench}{EmotionBench} & \green{global} & \green{holistic} & \red{deterministic} (P) & \citet{chenEmotionQueenBenchmarkEvaluating2024} \\
& motivation & \href{https://github.com/seacowx/OpenToM}{OpenToM} & \green{local} & \red{progressive} (D) & \red{deterministic} (P) & \citet{xuOpenToMComprehensiveBenchmark2024} \\
& & \href{https://github.com/skywalker023/fantom}{FANTOM} & \green{global} & \red{holistic} (P) & \red{deterministic} (P) & \citet{kimFANToMBenchmarkStresstesting2023} \\
& & \href{https://github.com/microsoft/AnthropomorphicIntelligence/blob/main/MotiveBench/README.md}{MOTIVEBENCH} & \green{global} & \red{holistic} (P) & \red{deterministic} (P) & \citet{yongMotiveBenchHowFar2025} \\
& & \href{https://github.com/thu-coai/CharacterBench}{CHARACTERBENCH} & \green{global} & \red{holistic} (P) & \red{deterministic} (P) & \citet{zhouCharacterBenchBenchmarkingCharacter} \\
\midrule
\multirow{2}{*}{\textbf{Social}} 
 & interaction type & --- & --- & --- & --- & --- \\
& connections & \href{https://github.com/kilian-group/phantom-wiki}{PhantomWiki} & \green{global} & \green{holistic} & \green{deterministic} & \citet{gongPhantomWikiOnDemandDatasets2025} \\
& relationship & \href{https://github.com/kwai/DialogBench}{DialogBench} & \green{global} & \green{holistic} & \red{deterministic} (P) & \citet{ouDialogBenchEvaluatingLLMs2024} \\
\midrule
\textbf{Event} & causality & --- & --- & --- & --- & --- \\
 & event & --- & --- & --- & --- & --- \\
 & schema & --- & --- & --- & --- & --- \\
\midrule
\multirow{5}{*}{\textbf{Plot}}
& topic & \href{https://github.com/ServiceNow/repliqa}{RepLiQA} & \green{global} & \green{holistic} & \red{deterministic} (C) & \citet{monteiroREPLIQAQuestionAnsweringDataset} \\
& plot & \href{https://github.com/kabirahuja2431/FlawedFictions}{FlawedFictions} & \green{global} & \red{progressive} (H) & \red{deterministic} (P) & \citet{ahujaFindingFlawedFictions2025a} \\
& plotline & \href{https://github.com/melaniesubbiah/storysumm}{StorySumm} & \green{global} & \green{holistic} & \green{consensus} & \citet{subbiahSTORYSUMMEvaluatingFaithfulness2025} \\
& moral & \href{https://github.com/agiresearch/MoralBench}{MoralBench} & \green{global} & \green{holistic} & \red{deterministic} (P) & \citet{jiMoralBenchMoralEvaluation} \\
& & \href{https://huggingface.co/datasets/cardiffnlp/Morables}{MORABLES} & \green{global} & \green{holistic} & \red{deterministic} (P) & \citet{marcuzzoMORABLESBenchmarkAssessing2025} \\
& obstacle & --- & --- & --- & --- & --- \\
& conflict & --- & --- & --- & --- & --- \\
& archetype & --- & --- & --- & --- & --- \\
\midrule
\textbf{Setting} & setting & \href{https://zenodo.org/records/16670471}{Dataset-GSS} & \green{global} & \green{holistic} & \red{deterministic} (C) & \citet{liPixelsPlacesSystematic2025} \\
 & location & --- & --- & --- & --- & --- \\
\midrule
\multirow{2}{*}{\textbf{Structure}} 
& plot arc & \href{https://github.com/cltl/EventStoryLine}{EventStoryLine} & \green{global} & \red{holistic} (P) & \red{deterministic} (C) & \citet{caselliEventStoryLineCorpus2017} \\
& & \href{https://github.com/PlusLabNLP/Narrative-Discourse}{Narrative-Discourse} & \green{global} & \red{holistic} (P) & \red{deterministic} (C) & \citet{tianAreLargeLanguage2024} \\
\midrule
\multirow{2}{*}{\textbf{Perspective}} 
& dialogue & \href{https://github.com/johndmendonca/soda_eval?tab=readme-ov-file}{SODA-EVAL} & \red{global} (L) & \red{holistic} (H) & \green{deterministic} & \citet{mendoncaSodaEvalOpenDomainDialogue2024} \\
& & \href{https://github.com/Neph0s/CoSER}{CoSER} & \red{global} (L) & \red{holistic} (H) & \green{deterministic} & \citet{wangCoSERCoordinatingLLMBased2025} \\
 & point of view & --- & --- & --- & --- & --- \\
 & focalization & --- & --- & --- & --- & --- \\
\midrule
\textbf{Style} & allusion & --- & --- & --- & --- & --- \\
 & figurative lang. & --- & --- & --- & --- & --- \\
 & imageability & --- & --- & --- & --- & --- \\
 & complexity & --- & --- & --- & --- & --- \\
 & evaluative lang. & --- & --- & --- & --- & --- \\
\midrule
\multirow{4}{*}{\textbf{Time}}
 & duration & --- & --- & --- & --- & --- \\
& order & \href{https://gitlab.ub.uni-bielefeld.de/s.kenneweg/TRaVelER}{TRaVelER} & \green{global} & \green{holistic} & \green{deterministic} & \citet{kennewegBenchmarkingAbilityLarge2024} \\
& & \href{https://huggingface.co/datasets/baharef/ToT}{ToT} & \green{global} & \green{holistic} & \green{deterministic} & \citet{fatemiTestTimeBenchmark2024} \\
& & \href{https://github.com/THU-KEG/MAVEN-ERE}{MAVEN-ERE} & \green{global} & \green{holistic} & \green{deterministic} & \citet{wangMAVENEREUnifiedLargescale2022} \\
& & \href{https://github.com/EternityYW/TRAM-Benchmark}{TRAM} & \green{global} & \green{holistic} & \green{deterministic} & \citet{wangTRAMBenchmarkingTemporal2024} \\
\midrule
\textbf{Revelation} & suspense & \href{https://github.com/zhaochen0110/conflictbank}{ConflictBank} & \green{global} & \red{holistic} (P) & \red{deterministic} (P) & \citet{suConflictBankBenchmarkEvaluating2024} \\
 & curiosity & --- & --- & --- & --- & --- \\
 & surprise & --- & --- & --- & --- & --- \\
\midrule
\multirow{4}{*}{\textbf{Paratext}} 
& genre & \href{https://github.com/uchidalab/book-dataset}{BookCover30} & \green{global} & \red{discrete} (H) & \red{deterministic} (C) & \citet{iwanaJudgingBookIts2017} \\
& author & \href{https://github.com/bit-ml/VeriDark}{VeriDark} & \green{global} & \red{holistic} (D) & \green{deterministic} & \citet{manolacheVeriDarkLargeScaleBenchmark2022} \\
& & \href{https://github.com/YashSaxena21/REASONS}{REASONS} & \green{global} & \green{discrete} & \green{deterministic} & \citet{saxenaAttributionScientificLiterature2025} \\
& date & \href{https://huggingface.co/datasets/hereldav/TimeAware}{TimeAware} & \green{global} & \green{discrete} & \green{deterministic} & \citet{herelTimeAwarenessLarge2025} \\
 & medium & --- & --- & --- & --- & --- \\
 & platform & --- & --- & --- & --- & --- \\
\midrule
\textbf{Motivation} & intent & \href{https://huggingface.co/datasets/APauli/Persuasive-Pairs}{Persuasive Pairs} & \green{global} & \green{holistic} & \red{deterministic} (P) & \citet{pauliMeasuringBenchmarkingLarge2025} \\
\bottomrule
\end{tabular}
\caption{Benchmark survey results showing existing benchmarks mapped to \textsc{NarraBench} taxonomy attributes. Green indicates match with desired criteria, red indicates mismatch, with initials in brackets indicating the preferred property according to our taxonomy.}
\label{tab:survey}
\end{table*}

\end{document}